\documentclass[lettersize,journal]{IEEEtran}
\usepackage{amsmath,amsfonts}
\usepackage{algorithmic}
\usepackage{algorithm}
\usepackage{array}
\usepackage[caption=false,font=normalsize,labelfont=sf,textfont=sf]{subfig}
\usepackage{textcomp}
\usepackage{stfloats}
\usepackage{url}
\usepackage{verbatim}
\usepackage{graphicx}
\usepackage{cite}
\usepackage{booktabs}
\usepackage{enumerate}
\usepackage{epsfig}
\usepackage{amssymb}
\usepackage{amsthm}
\usepackage{bm}
\usepackage[pagebackref,breaklinks,colorlinks]{hyperref}

\newcommand{\tabincell}[2]{\begin{tabular}{@{}#1@{}}#2\end{tabular}}
\usepackage{diagbox}
\usepackage{multirow}
\begin{document}

\title{HIH: Towards More Accurate Face Alignment \\ via Heatmap in Heatmap}

\author{\IEEEauthorblockN{Xing Lan\IEEEauthorrefmark{4}\IEEEauthorrefmark{1},
Qinghao Hu\IEEEauthorrefmark{2}\IEEEauthorrefmark{1},
Qiang Chen\IEEEauthorrefmark{3},
Jian Xue\IEEEauthorrefmark{4},
Jian Cheng\IEEEauthorrefmark{4}\IEEEauthorrefmark{2}
}

\IEEEauthorblockA{\IEEEauthorrefmark{4}
University of Chinese Academy of Sciences, Beijing, 100049 China}

\IEEEauthorblockA{\IEEEauthorrefmark{2}
Institute of Automation, Chinese Academy of Sciences, Beijing, 100190 China}

\IEEEauthorblockA{\IEEEauthorrefmark{3}
Baidu Inc., Beijing, 100085 China}

\thanks{Xing Lan and Jian Cheng are with the National Laboratory of Pattern Recognition, 
Institute of Automation Chinese Academy of Sciences 
and University of Chinese Academy of Sciences, Beijing, China. 
Qinghao Hu is with National Laboratory of Pattern Recognition, 
Institute of Automation Chinese Academy of Sciences.
Qiang Chen is with Baidu Inc., Beijing, China.
Jian Xue is with University of Chinese Academy of Sciences, Beijing, China. 
(e-mail: xinglan97@gmail.com; huqinghao2014@ia.ac.cn; chenqiang13@baidu.com; xuejian@ucas.ac.cn; jcheng@nlpr.ia.ac.cn).
\IEEEauthorrefmark{1} indicates equal contribution.
Corresponding author: Jian Cheng (email: jcheng@nlpr.ia.ac.cn).}}
        
\markboth{Journal of \LaTeX\ Class Files,~Vol.~14, No.~8, August~2021}%
{Lan \MakeLowercase{\textit{et al.}}: HIH: Towards More Accurate Face Alignment via Heatmap in Heatmap}


\maketitle

\begin{figure*}[!tbhp]
  \begin{center}
  \includegraphics[width=0.98\linewidth]{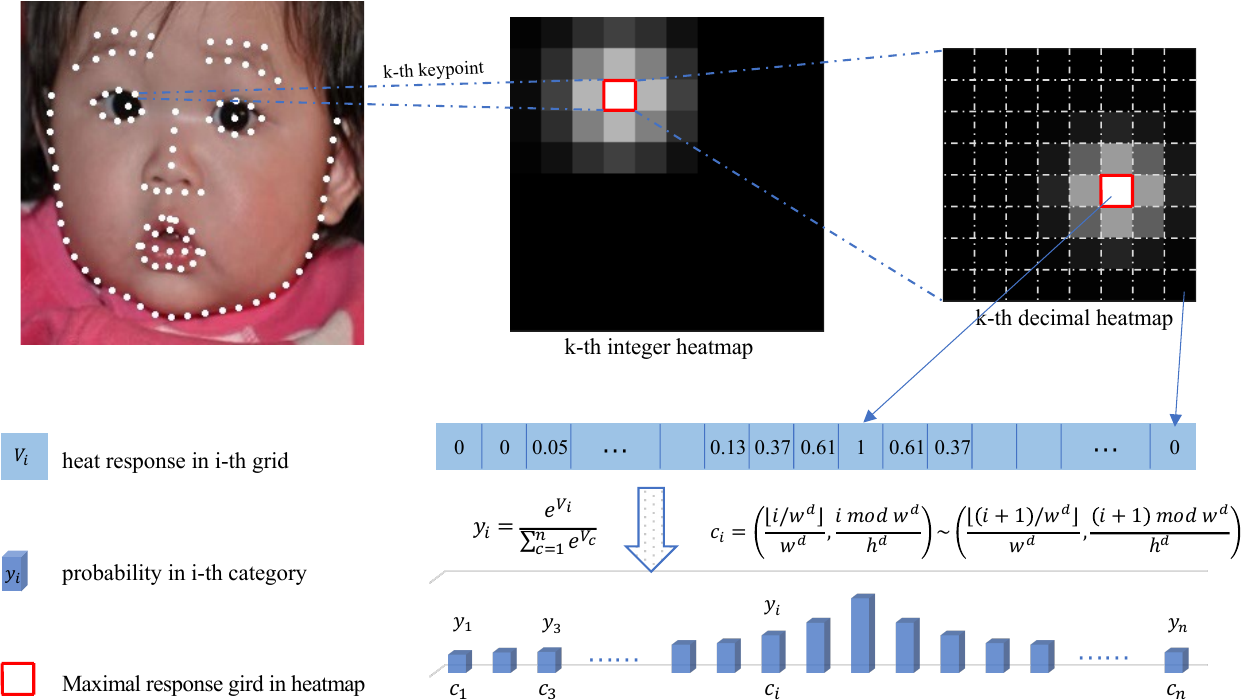}
  \end{center}
  \caption{HIH representation of the k-th keypoint. HIH generates two sets of heatmaps, integer heatmap and decimal heatmap, which stand for integer coordinate and subpixel coordinate.
  The light blue strip is flattened from the decimal heatmap, and the values are the corresponding heat response in the heatmap.
  The blue cuboid represents an interval category, and its height stands for its probability.
  The number of girds is the product of the decimal heatmap's height and width. 
  HIH regards the response in the i-th grid as the probability in the i-th category, and each category $c_i$ represents the corresponding subpixel-coordinate interval. 
    }
  \label{tab:quantization error}
  \end{figure*}

\begin{abstract}
Heatmap-based regression overcomes the lack of spatial and contextual information of direct coordinate regression, and has revolutionized the task of face alignment.
Yet it suffers from quantization errors caused by neglecting subpixel coordinates in image resizing and network downsampling.
In this paper, we first quantitatively analyze the quantization error on benchmarks, which accounts for more than 1/3 of the whole prediction errors for state-of-the-art methods.
To tackle this problem, we propose a novel Heatmap In Heatmap(HIH) representation and a coordinate soft-classification (CSC) method, which are seamlessly integrated into the classic hourglass network.
The HIH representation utilizes nested heatmaps to jointly represent the coordinate label:
one heatmap called integer heatmap stands for the integer coordinate, and the other heatmap named decimal heatmap represents the subpixel coordinate.
The range of a decimal heatmap makes up one pixel in the corresponding integer heatmap.
Besides, we transfer the offset regression problem to an interval classification task, and CSC regards the confidence of the pixel as the probability of the interval. 
Meanwhile, CSC applying the distribution loss leverage the soft labels generated from the Gaussian distribution function to guide the offset heatmap training, which makes it easier to learn the distribution of coordinate offsets.
Extensive experiments on challenging benchmark datasets demonstrate that our HIH can achieve state-of-the-art results. 
In particular, our HIH reaches 4.08 NME (Normalized Mean Error) on WFLW, and 3.21 on COFW,
which exceeds previous methods by a significant margin.

\end{abstract}


\section{Introduction}

Face alignment, or facial landmark detection, refers to detecting a set of predefined landmarks on the human face, which is a fundamental step for many facial tasks\cite{liu2017sphereface,khan2017synergy,feng2018joint}.
Heatmap-based methods\cite{kowalski2017deep,newell2016stacked,deng2019joint,yang2017stacked} use the heatmap as label representation to encode the coordinate of facial landmark labels so that the supervised learning loss can be quantified.
The heatmap is generated by the 2-dimension Gaussian distribution density function, and it is characterized by giving spatial support around the ground-truth location, 
considering not only the contextual clues but also the inherent target position ambiguity \cite{zhang2020distribution}.
In this way, this label representation method effectively alleviates the model overfitting phenomenon during the training procedure.

However, there is a significant obstacle to the heatmap representation, which is reflected in the following.
Since the computational cost is a quadratic function of the input image resolution, 
downsampling layers are introduced into deep networks to ease the computation burden, 
and it causes the size of network output is usually smaller than the input image.
For heatmap-based methods,  the prediction of a landmark is considered as the location (integer type) with maximal activation, which is required to remap back to the original coordinate space.
Thus, the prediction is the sub-optimal coordinate, and the quantization error is introduced during the resolution reduction, which drops the subpixel coordinate and causes precision decreases \cite{papandreou2017towards,luvizon20182d,jin2020pixel}.

How much impact does the quantization error cause on the final results?
We are the first to conduct quantitative analysis on benchmarks. 
Under the common setting \cite{HRNET,kumar2020luvli}, from 256x256 input to 64x64 heatmap, the bottleneck caused by quantization error is shown in Fig.\ref{fig:bottleneck}.
The normalized mean error (NME) generated by quantization error is 1.285 on WFLW\cite{LABWFLW}, 1.111 on 300W\cite{300W}, even larger than 1/3 of the state-of-the-art methods caused NME.
Thus, how to eliminate the quantization effect is significant in face alignment.

At present, there are some solutions to alleviate the error, which are mainly divided into two categories: 
{\bf 1.} leveraging soft-argmax\cite{luvizon20182d,bulat2021subpixel}, vote\cite{papandreou2017towards}, mean\cite{kumar2020luvli}, or Gaussian operation\cite{zhang2020distribution,wan2020robust} based on the adjacent coordinates\cite{HRNET,newell2016stacked,tai2019towards};
{\bf 2.} using an extra branch of network to learn the subpixel coordinate, which is expressed in the form of offset value \cite{zhang2020robust,xia2022sparse,jin2020pixel} and offset map \cite{law2018cornernet,duan2019centernet,zhou2019bottom}.

However, the former only has coarse-grained supervision while lacking fine-grained supervision, and the post-processing operation of most methods\cite{HRNET,luvizon20182d}  is based on manual design rather than learnable modules.
And when the resolution of the heatmap is low, the points on the heatmap are sparse, and some spatial distribution details are lost, which makes it difficult to estimate the coordinates\cite{yin2020attentive}.
For the latter, the offset value representation has not taken full advantage of spatial support, 
and the offset map representation is not suitable for dense point estimation tasks because the location of dense points is easy to conflict.

To address the above issues, we design a novel representation to refine the subpixel coordinate, called \textbf{H}eatmap \textbf{I}n \textbf{H}eatmap (HIH), which belongs to the offset-branch category but with a novel design.
HIH proposes two categories of heatmaps as label representations to encode the coordinate of landmark labels.
One keeps consistent with the original to represent integer value, called integer heatmap.
The other one represents the remaining subpixel value of location, named decimal heatmap. 
The range of one decimal heatmap represents one pixel in the corresponding integer heatmap, so the design is named \emph{Heatmap in Heatmap}.
Both heatmaps take the location with maximal activation as the coordinate result, 
the location decoding from the integer heatmaps is the sub-optimal coordinates, 
and the final prediction is represented by the sum of the integer coordinate and the normalized fractional coordinate. 
Compared with previous works, HIH takes full advantage of spatial support and contextual information.
Moreover, the fine-grained supervision of subpixel coordinates is in a classification way to constrain the decimal heatmaps.

Inspired by the characteristic (discussed in subsec.\ref{loss_discuss}) of the $\arg \max$ function and the limited resolution of the decimal heatmap, we regard the prediction of the decimal heatmap as a classification problem to solve.
In one decimal heatmap, we use each location of pixels to represent one category and regard the confidence of the pixel as the probability of the interval category.
Therefore, the number of categories is the product of the height and width of the decimal heatmap, and all intervals make up one pixel in the integer heatmap.
Besides, we design a novel loss function, called \textbf{C}oordinate \textbf{S}oft-\textbf{C}lassification Loss ($\mathcal{L}_{csc}$), which effectively leverage the soft labels generated by 2-dimension gaussian distribution density function.



To verify the feasibility of our method, we take three implementation designs of structure (tiny, small, and base).
Extensive experiments on various benchmarks are conducted to demonstrate the superior performance of HIH to other solutions. 
Specifically, the NME reaches \textbf{4.08} on WFLW \cite{LABWFLW}, which outperforms SOTA with a significant margin.
The code is available at the project website\footnote{https://github.com/starhiking/HeatmapInHeatmap}.
The main contributions in this work are summarized as following.

1. To the best of our knowledge, this is the first study to quantitatively analyze the quantization error on benchmarks. The bottleneck surprisingly accounts for more than 1/3 as SOTA caused.
To alleviate the problem, we propose a heatmap-in-heatmap representation, and redesign its encoding and decoding approaches. 
HIH also uses the heatmap to stand for the subpixel coordinate, which takes full advantage of spatial support and essentially eliminates the quantization error introduced by the heatmap discretization process.

2. We further propose to solve the prediction of the decimal heatmap with the idea of classification. 
Besides, we introduce a novel loss function that leverages the features of argmax function and gaussian distribution, which makes it possible to learn the relative distribution. 

3. Detailed experiments demonstrate the feasibility of HIH in compensating quantization error as well as superiority over other solutions, and $\mathcal{L}_{csc}$ further boosts the performance of HIH. 
Our approach outperforms the state-of-the-art algorithms by a significant margin on various benchmarks.

Different from our prior work\cite{lan2021revisting} focus on the representation design of coordinates,
we transform the subpixel regression problem into an interval classification problem.
Moreover, we design a seamless loss function and further improve the performance.
We also conduct more detailed experiments and ablation studies, as well as implement more comprehensive metrics to analyze the experiments.

\section{Related Work}
Early methods were based on active shape models (ASMs) \cite{cootes1992active,Timothy1995Active,milborrow2008locating,romdhani1999multi} and active appearance models (AAMs) \cite{cootes2001active,sauer2011accurate,matthews2004active,kahraman2007active,cootes2000view,tzimiropoulos2013optimization}.
Later constrained local models (CLMs) \cite{cristinacce2008automatic,zhu2016unconstrained,tzimiropoulos2015project,xiong2013supervised} and the subsequent cascaded regression methods \cite{asthana2014incremental,dollar2010cascaded,tuzel2016robust,feng2015cascaded} promote the accuracy of face alignment.
Recently, deep learning methods have dominated the best performance in this area, which mainly fall into two categories: direct coordinate regression  and heatmap-based regression.

\subsection{Coordinate Regression Models.} 
Coordinate regression methods \cite{sun2013deep,zhang2014facial,feng2018wing,zhang2016joint,dong2018style} can be used to directly map 
a facial image to the target landmark coordinates. In the architecture of deep learning methods,
the features of input image are usually extracted by a convolution neural network (CNN) and
then mapped to coordinates through fully-connected layers.
Diverse cascaded networks \cite{trigeorgis2016mnemonic} 
and recurrent networks\cite{xiao2016robust} are proposed to achieve face alignment with multi stages. 
To further improve the robustness, some methods\cite{wu2017leveraging,smith2014collaborative,zhu2014transferring,zhang2015leveraging,lan2020atf} leverage other datasets in across training to learn more data distribution.
3FabRec \cite{browatzki20203fabrec} and \cite{dong2020supervision} leverage the unlabeled dataset to train the model.
Wing loss \cite{feng2018wing} was proposed as a new loss function for landmark detection, which can obtain robust performance against widely used L2 loss.
SAN \cite{dong2018style} and AVS.SAN \cite{qian2019aggregation} accompanies the original face images with style aggregated ones to train the landmark detector.
Recently, state-of-the-art works employ the structure information of face as the prior knowledge for better performance. 
SDFL \cite{lin2021structure} and Li et al. \cite{li2020structured} model the interaction between landmarks by a graph convolutional network. 


\subsection{Heatmap-based Face Alignment}
Heatmap-based methods \cite{kowalski2017deep,deng2019joint,yang2017stacked,jin2020pixel} show better performance than direct coordinate regression due to their spatial support.
Heatmap is characterized by giving spatial support around the ground-truth location, which effectively reduces the model overfitting risk in training.
Heatmap regression methods are used to indirectly map the input image to the probability heatmaps, each of which represents the respective probability of a landmark location.
In the inference stage, the location with the highest response on each heatmap indicates the corresponding landmark.
DAN \cite{kowalski2017deep} is the first method that combines heatmap with landmarks regression. 
LAB \cite{LABWFLW} and Awing \cite{wang2019adaptive} use additional boundary lines as the geometric structure of a face image to help facial landmark localization.
HRNet \cite{HRNET} maintains multi-resolution representations in parallel and exchanges information between these streams 
to obtain a final representation with great semantics and precise locations.
LUVLI \cite{kumar2020luvli} first introduces the concept of parametric uncertainty estimation as well as considers the visibility likelihood.
ADNet \cite{huang2021adnet} further regress the coordinate from anisotropic direction loss and anisotropic attention module.
Recently, SLPT \cite{xia2022sparse} leverages the coarse-to-fine way and Transformer \cite{carion2020end,vaswani2017attention} structure to refine the coordinate.

\subsection{Quantization Error in Pixel-level Tasks}

In general, due to the high computation demand for maintaining high resolution, the shape of the output heatmap is smaller than the original input image by using downsampling operations.
As a result, these downsampling operations bring in  quantization error since the heatmap’s argmax is only determined to the nearest pixel, while the subpixel value is neglected.
Specifically, in face alignment networks \cite{HRNET,kumar2020luvli,yang2017stacked}, the shape of the input image is 256 pixels for width and height, while the output heatmap is 64 pixels.
In other words, no matter how effective the above methods are, there will be the quantization error from 4 times down-sample operations that influence the accuracy.

In the literature on face alignment, there are the following works proposed to eliminate the error.
FHR \cite{tai2019towards} leverages three heatmap coordinates and corresponding confidence to jointly predict the subpixel parts via a Gaussian function.
LUVLI \cite{kumar2020luvli} chooses a mean estimator for the heatmap, which is differentiable and enables sub-pixel accuracy.
Yin et al. \cite{yin2020attentive} designs a regress method based on 1D heatmaps which represent the marginal distribution on each axis.
MHHN \cite{wan2020robust} proposes a subpixel detection loss and subpixel detection technology to achieve high-precision face alignment via a heatmap subpixel regression model.
SHN-GCN \cite{zhang2020robust} combines feature fusion strategy and attention mechanism to capture spatial representations and globally refines the landmarks localization.
Bulat et al. \cite{bulat2021subpixel} takes advantage of a local soft-argmax way to redesign the encoding and decoding method, which can leverage the underlying continuous distribution.
SLPT \cite{xia2022sparse} learns the adaptive inherent relation to refine the subpixel coordinate in multiple stages.

In the areas of pose estimation and object detection, the following methods also have achieved impressive results in their area.
G-RMI \cite{papandreou2017towards} also takes the mask into consideration and uses all the surrounding response to vote for the weighted value.
Hourglass \cite{newell2016stacked} first identifies the coordinates of the maximal and second maximal activation, and then uses the position and gradient to fine-tune the maximum response coordinate.
DARK \cite{zhang2020distribution} first smooths the irregular heatmap by Gaussian smoothing, then expands Taylor's formula to fit the derivative of the Gaussian function, and finally calculate the position of the maximum.
Luvizon et al.\cite{luvizon20182d} carries out a global soft-argmax operation and then takes advantage of joint spatial mean to get final coordinates.
CornerNet\cite{law2018cornernet}, CenterNet\cite{duan2019centernet} and ExtremeNet \cite{zhou2019bottom} propose to use a new branch to regress the offset caused by quantization in object detection, which will produce no bias in encoding.

However, the methods based on the near points only have coarse-grained supervision while lacking fine-grained supervision, 
and the post-processing operation of most methods is based on manual design rather than learnable.
And when the resolution of the heatmap is low, the points on the heatmap are sparse, and some spatial distribution details are lost, which makes it difficult to estimate the coordinates\cite{yin2020attentive}.
The offset value representation \cite{xia2022sparse} has not taken full advantage of spatial support, 
and the offset map representation is not suitable for dense point estimation tasks because the location of dense points is easy to conflict.
Our designed representation takes full advantage of spatial support and contextual information. 
Moreover, the fine-grained supervision of subpixel coordinates is with classification way to constrain the decimal heatmaps. 
Compared with \cite{law2018cornernet}, the representation of each point corresponds to an integer heatmap and a decimal heatmap. 
There will be no positional conflict between points, which is suitable for dense points localization.

\begin{figure*}[tbh]
  \centering
  \newpage
  \includegraphics[width=\linewidth]{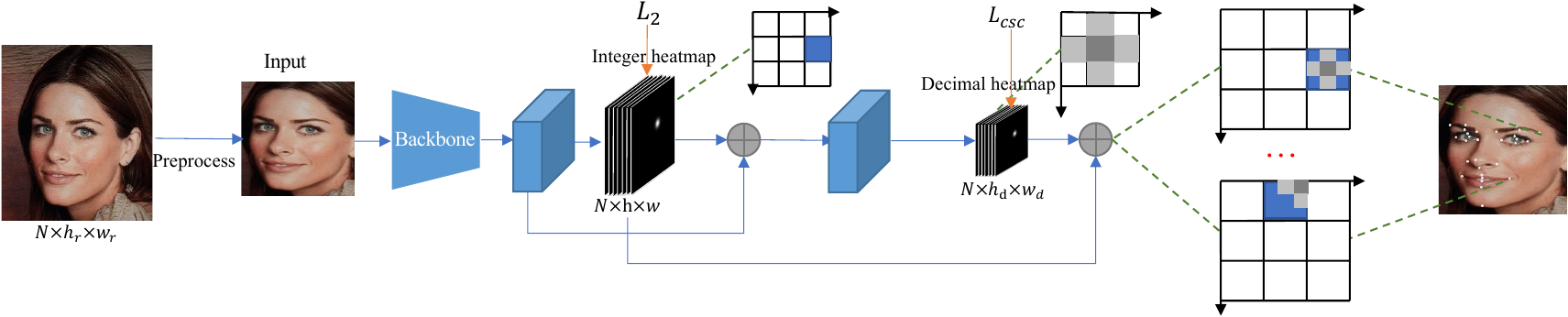}
  \caption{The overall pipeline of training process.
  There exist two kinds of heatmaps that jointly represent coordinate labels: integer heatmap identifies integer coordinate and decimal heatmap refines its remaining fractional location.
  The blue grid represents the location with the maximal response in the integer heatmap. 
  And dark gray, light gray, and white grids correspond to the soft labels with different probabilities in the decimal heatmap, respectively.
  }
  \label{fig:method}
\end{figure*}

\section{Methodology}
\subsection{Overview}
\begin{figure}[!tbh]
  \centering
  \newpage
  \includegraphics[width=0.58\linewidth]{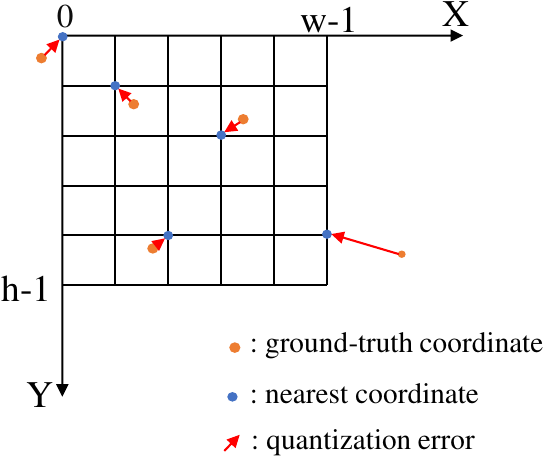}
  \caption{Illustration of quantization error in heatmap.
  The yellow points are ground-truth coordinates, 
  and the blue points are nearest coordinates after $round$ operation.
  Each red arrow is the distance between them, which represents the quantization error introduced by mentioned operation.
  }
  \label{fig:error}
\end{figure}

As shown in Fig.\ref{fig:error}, the quantization error is introduced by neglecting the subpixel coordinate when image resizing and network downsampling. 
One pixel in the original image corresponds to less one-pixel space in the heatmap after image-resizing and downsampling. 
Yet the heatmaps only produce integer values that can not represent the decimal (or fractional) coordinates. 

To eliminate the effect introduced by quantization error, we propose a novel \emph{Heatmap in Heatmap} (HIH) method, 
which takes two categories of heatmaps to jointly represent coordinate values.
As shown in Fig.\ref{fig:method}, one identifies the integer location, named \emph{integer heatmap}, and the other locates its subpixel location, called \emph{decimal heatmap}.
At the integer-heatmap coordinate space, integer heatmap represents the integer part from 0 to its resolution,
and the decimal heatmap represents the remaining offset from 0 to 1.
In other words, the whole decimal heatmap represents one pixel in the corresponding integer heatmap, 
and those subpixel coordinates, which are less than one pixel and ignored in the integer heatmap, will be represented by the decimal heatmap. 
Proposition \ref{props} demonstrates that our method has a smaller coordinate representation precision\footnote{Here coordinate representation precision means the smallest change that can be represented by the heatmap. Generally speaking, the quantization error mostly comes from a large representation precision.} 
than one pixel in the original image under a loose condition, which means that our method can reduce quantization errors by decreasing coordinate representation precision. 
For example, given a $512\times 512$ image, after resizing and downsampling, the landmarks are represented by a set of $64\times64$ integer heatmaps and $8\times8$ decimal heatmaps. 
The representation precision of our method is one pixel in the original image, which means that our method can represent any pixel in the original image by the combination of integer and decimal heatmap. 
Compared with previous works, the decimal heatmap offers enough spatial support and fine-grained supervision.

Inspired by the definition of argmax function (common decode approach) and the limited resolution of decimal heatmaps, we further propose to learn the relative orders of decimal heatmap values by using classification instead of regression. 
Meanwhile, we introduce a soft-label-based classification loss (CSC) instead of the common MSE loss to constrain the influence of mutual interactions between pixel values.
In the next subsections, we give details of representation design and loss definition.


\newtheorem{prop}{Proposition}[section]
\begin{prop} \label{props}
  Given one set of integer heatmap and decimal heatmap, our HIH has a smaller coordinate representation precision than one pixel in the original image if the resolution of the original image is below or equal to the product of the integer heatmap's resolution and decimal heatmap's one. 
\end{prop}

\begin{proof}
Suppose the resolution of original images is $h_r \times w_r$, and resized to the input size of CNNs during image pre-processing. 
Let the resolution of integer heatmap and decimal heatmap  be $h\times w$  and $h^d \times w^d$, respectively. 
In HIH, we represent one pixel in the integer heatmap by the whole decimal heatmap.
Thus one pixel in the decimal heatmap only takes ($\frac{1}{h^d}$,$\frac{1}{w^d}$) pixel of the integer heatmap. 
Besides, all the coordinates in the integer heatmap will be scaled by ($\frac{h^r}{h}$,$\frac{w^r}{w}$) times to get the coordinates in the original images. 
As a result, one pixel in decimal heatmap only represents ($\frac{h^r}{h*h^d}$,$\frac{w^r}{w*w^d}$) pixels of original images. 
If $h^r\le h*h^d$ and  $w^r\le w*w^d$, our HIH has a smaller coordinate precision than one pixel  since $\frac{h^r}{h*h^d}\le 1$ and $\frac{w^r}{w*w^d} \le 1$.
\end{proof}

\subsection{Encoding and Decoding}
Different from the near points-based methods\cite{luvizon20182d,kumar2020luvli}, we propose an extra decimal heatmap to represent the subpixel part as the remaining offset.
Meanwhile, the original heatmap still represents the sub-optimal coordinate (integer value).
The whole range of decimal heatmap represents one pixel in the integer heatmap.
HIH's integer and decimal location are represented by the 2-dimensional Gaussian distribution density function but with different settings (sigma, resolution, $etc.$).
Specifically, given a task for detecting $K$ facial landmarks, the input image is transformed into integer heatmaps ($K,h,w$) and decimal heatmaps ($K,h^d,w^d$) during model computation.

\textbf{Encoding design.}
The original coordinate labels involve two heatmaps encoding.
Suppose integer heatmap is denoted as $H^I$, we first downsample the coordinate label by corresponding times and take the floor operation to get the integer value.
Then we utilize the 2-dimensional Gaussian distribution density function to generate the integer heatmap as following.

\begin{equation}
  H^I(x,y) = 
  \begin{cases}
    \exp(- \frac{(x - \lfloor \frac{u}{n} \rfloor )^2 + (y - \lfloor \frac{v}{n} \rfloor )^2 }{2\sigma^2}) & d \leq 3\sigma \\
    0 & \text{otherwise}
  \end{cases}
\end{equation}

where $n$ is the downsampling factor, $(u,v)$ is the ground-truth label for the original image, $(\lfloor \frac{u}{n} \rfloor, \lfloor \frac{v}{n} \rfloor)$ denotes the groud-truth coordinate on the integer-heatmap coordinate space.
$\sigma$ is the standard deviation, $(x,y)$ stands for the integer coordinates in the heatmap, and $d = ||(x,y)-(\frac{u}{n},\frac{v}{n})||_{2}$.

Suppose decimal heatmap is denoted as $H^D$, we first get the subpixel coordinate according to the suboptimal and ground-truth coordinates.
Because the fractional value ranges from 0 to 1, and represents the coordinates of maximal activation in the decimal heatmap, 
it is required to be enlarged by the resolution ratio.
And the final integer coordinate $c^D$ on decimal heatmap is with the round operation as following.

\begin{equation}
    \label{eq:hih}
  \begin{aligned}
  c^D = & (\lfloor (\frac{u}{n} - \lfloor \frac{u}{n} \rfloor)*w^d \rceil, \lfloor (\frac{v}{n} - \lfloor \frac{v}{n} \rfloor)*h^d \rceil) \\
\end{aligned}
\end{equation}

where $u,v,n$ are the same meaning as before, $(h^d,w^d)$ is the resolution of decimal heatmaps.
And we also utilize the same distribution density function to produce the decimal heatmaps as following.

\begin{equation}
  \label{Gaussian distribution}
\begin{aligned}
   H^D(x,y) = &
    \begin{cases}
       \exp(- \frac{ (x- c_x^D) )^2 +  (y- c_y^D )^2  }{2\sigma^2}  ) & d \leq 3\sigma \\
       0 & \text{otherwise}
    \end{cases}
\end{aligned}
\end{equation}

where $x,y,d,\sigma$ are the similar meaning in integer heatmaps as before while $\sigma$ is the super parameter and with different settings in decimal heatmaps.
By using the designed encoding way, there are the nested heatmaps generated: the integer heatmaps $H^I$ with $K\times h\times w$ shape size and decimal heatmaps $H^D$ with $K\times h^d\times w^d$ size. 

\textbf{Decoding design.}
For our proposed representation, the integer heatmaps are required to be decoded to the suboptimal coordinates, 
and the decimal heatmaps are also decoded to the subpixel coordinates which is required to be narrowed to the stand coordinate space.
Due to the offset representation, the integer heatmap uses the maximal response location as the integer value. 
And because the resolution of decimal heatmaps is low \cite{yin2020attentive}, decoding decimal heatmap still uses the argmax function.
Suppose the predicted integer heatmaps $\hat{H^I}$ and decimal heatmap $\hat{H^D}$, which are both composed of the location's activation.
The final normalized coordinate $\bm{\hat{P}}$ is predicted by

\begin{equation}
  \begin{aligned}
    \bm{\hat{P}} &=  (\hat{c}^I +  \hat{c}^D / r^D ) / r^I \\
                   &=  \frac{\arg \max \hat{H}^I}{(w,h)}  + \frac{\arg \max \hat{H}^D}{(w\cdot w^d,h\cdot h^d)}  
  \end{aligned}
\end{equation}


 
 

where $r^I$ and $r^D$ represent the resolution size of integer heatmap and decimal heatmap, $\hat{c}^I$ and $\hat{c}^D$ represent the predicted integer coordinate and fractional coordinate, respectively.

The resolution of the decimal heatmap represents the ideal precision of coordinate recovery to solve the quantization error.
Take the integer heatmap resolution of 64 $\times$ 64 and the decimal heatmap resolution of 8 $\times$ 8 as an example, the joint representation can unbiasedly encode and decode any pixel location under $512 \times 512$ coordinate space.
As shown in the Ideal item of Tab.\ref{ablation_resolution}, HIH obviously decreases the quantization error, and the impact introduced by the error is now reduced to 4\% (the previous impact is larger than 1/3) compared to state-of-the-art methods caused NME, which is no longer a bottleneck.


\subsection{Coordinate Soft-Classification Loss}
\label{loss_discuss}



In HIH design, one pixel value in a decimal heatmap stands for the response of the corresponding subpixel coordinate. 
The closer the response value is to 1, the more likely pixel's location is the fractional coordinate of the keypoint, and vice versa.
It is worth noting that the argmax or soft-argmax function pays more attention to the relative order (i.e., which one is the largest pixel) than the pixel value itself (i.e. how large is the pixel value), thus the relative ranking of pixel points in the heatmap is more critical.  
The decoding step in our method utilizes $\arg \max$ function to obtain the location of the maximum response among all pixels, which is required to find the location of the largest response instead of the response value. 
Yet the MSE loss, which is commonly used in heatmap-based models, only regresses the predicted value to be close to the ground-truth value. Still, it cannot constrain the magnitude relationship among the predicted values, especially when the corresponding ground-truth values are very close.
To cure this problem, we propose to learn the relative orders of decimal heatmap values by using classification instead of regression, and introduce a soft-label-based classification loss to guide the distribution of decimal heatmaps.
The limited resolution of the decimal heatmap gives us the chance to use the classification scheme because a large resolution brings too many categories which are challenging to train.  

In this paper, we regard the prediction of decimal heatmap as a classification problem to solve, and we introduce a novel loss function, called \textbf{C}oordinate \textbf{S}oft-\textbf{C}lassification Loss ($\mathcal{L}_{csc}$). 
Specifically, given $k$-th decimal heatmap $\hat{H}^D_k$ with size $h^d\times w^d$, we classify the pixels to $N_d=h^d*w^d$ categories, 
and each category represents the relative offset to the ground-truth landmark. 
The response value of the pixel represents the probability of each category.
In this way, the location with the maximal probability represents the location of $k$-th keypoint.
Common classification tasks use the one-hot labels as their categories differ a lot from each other semantically. 
In the decimal heatmaps, pixels near the largest response also have a high likelihood to be the keypoint. 
So we encode the decimal heatmap with Eq.\ref{Gaussian distribution} as described before, and we use the probability of decimal maps, which lies in the interval of $(0,1]$, as the soft labels for classification.
We take $\bm{q}$ to represent the pixel value of prediction $\hat{H^d}$ in the predict decimal heatmap, and $\bm{p}$ corresponds to the pixel value in ground-truth decimal heatmap $H_d$. 
We compute the soft-label based classification loss by using $KL-$divergence, and our $\mathcal{L}_{csc}$ is computed as following:
\begin{equation}
  \begin{aligned}  
    \mathcal{L}_{csc} &=KL(\bm{{p}^\star },\bm{{q}^\star }) \\
                      &=\frac{1}{K\cdot h^d \cdot w^d}\sum_{k}^K\sum_{i}^{h^d}\sum_{j}^{w^d} {\bm{{p}^\star }_{\left(k,i,j\right)}} \cdot \log \frac{\bm{{p}^\star }_{\left(k,i,j\right)}}{\bm{{q}^\star }_{\left(k,i,j\right)}}
  \end{aligned}  
\end{equation}

where ${\bm{{q}^\star }}$ = $softmax(\bm{q})$ and ${\bm{{p}^\star }}$ = $softmax(\bm{p})$, and 
$\bm{{p}^\star}_{\left(k,i,j\right)}$ represents the ground-truth value with location $\left( i,j\right)$ in $k$-th decimal heatmap after softmax operation. 
Therefore, the total loss $\mathfrak{L}$ is composed of the two parts, denoted as following. 

\begin{equation}
  \begin{aligned}
    \mathfrak{L} &= \mathcal{L}_{mse}+ \alpha \mathcal{L}_{csc} \\
                 &=  \frac{1}{K} \sum_{k}^K  (\frac{1}{h\cdot w} \sum_{i}^{h}\sum_{j}^{w} ||H^I(k,i,j) - \hat{H}^I(k,i,j)||_2^2  \\
                 &~~~~~~~~~~ + \frac{\alpha}{h^d \cdot w^d} \sum_{i}^{h^d}\sum_{j}^{w^d} {\bm{{p}^\star }_{\left(k,i,j\right)}} \cdot \log \frac{\bm{{p}^\star }_{\left(k,i,j\right)}}{\bm{{q}^\star }_{\left(k,i,j\right)}} )
  \end{aligned}
\end{equation}

Following previous practice\cite{HRNET}, here we take MSE loss to constrain the integer heatmap regression, $h$ and $w$ are the corresponding height and width.
$\alpha$ is the super-parameter for loss balance,  $H^I$ represents the ground-truth integer heatmap, and $\hat{H}^I$ corresponds to its prediction.

\subsection{Architecture design}
As shown in Fig.\ref{fig:method}, the whole network consists of three components: the backbone, integer part and  offset part.
Following previous works\cite{wang2019adaptive,LABWFLW,deng2019joint}, our backbone is based on the stacked Hourglass (HG) architecture\cite{newell2016stacked}.
For the last HG, the output heatmap (integer heatmap) is supervised with ground truth.
And the offset part uses the fusion of the backbone and integer heatmap as input,
We utilize three different weight implementations to verify the effectiveness of HIH.
For simplicity, we use the symbols $\text{HIH}_T$, $\text{HIH}_S$ and $\text{HIH}_B$ to represent the tiny, small, and base networks with 1, 2, and 4 stacked HGs respectively. 

To keep the similar computation as before solutions, we use  the above fusion input directly in the remaining part.
The first layer is a convolution whose stride equals 1, followed by a batch normalization layer and ReLU activation. 
Then the max-pooling layer reduces computation complexity by down-sampling feature maps.
The subsequent operation is a list of convolution blocks to learn spatial information, which uses the basic block of resnet\cite{he2016deep}.
In the above basic block, we set the stride to 2 in the first convolution, and to increase the receptive field quickly, we modify the kernel size of the second convolution to 3.
Inspired by HRNet, the final regression head comprises one convolution similar to the first layer and another single convolution.



\subsection{Difference between WOM and HIH}
\label{difference_WOM}

The solutions with offset map (WOM)\cite{law2018cornernet} and HIH both leverage maps to compensate for the quantization error, which is worth reiterating their differences.
WOM utilizes two maps $\left( 2 \times h \times w \right)$ to represent the offset, one map encodes x-offset value and the other represents y-offset.
It takes the integer value as the location of offset maps, and sets the x-offset and y-offset values on maps, respectively.
And our HIH utilizes a set of maps $\left( K \times h^d \times w^d \right)$ to encode the offset, which leverages the heatmap with its feature of spatial support.
HIH takes the offset value as locations as offset maps, and generates values by 2-dimension gaussian distribution function on decimal heatmaps, which decouples decimal and integer values.

WOM has not taken spatial information in the map and is deeply coupled with integer values, which is not suitable for dense point estimation tasks, as shown in Fig.\ref{wom_difference}.
Moreover, face alignment only detects a single point and cannot implicitly express the length and width information, which poses training more difficult.
Our experiments in subsection \ref{sec:comparison_solutions}  demonstrate that the combination with WOM has achieved slight improvement in face alignment.

\begin{table*}[tbh]
  \begin{center}
  \begin{tabular}{l|c|c|c|c|ccccccc}
  \hline
  \multirow{2}{*}{Method}&\multirow{2}{*}{Backbone}&\multirow{2}{*}{Params}&\multirow{2}{*}{Flops}&\multirow{2}{*}{Year}&\multicolumn{7}{c|}{WFLW}\\ \cline{6-12}
   &&&&&Fullset & \tabincell{c}{Pose} & \tabincell{c}{Exp.} & \tabincell{c}{Ill.} & \tabincell{c}{Mu.} & \tabincell{c}{Occ.} & \tabincell{c}{Blur} \\
  \hline\hline
  LAB \cite{LABWFLW} & 4 HG & 12.26M & 18.96G & $\text{CVPR}_{2018}$                 & 5.27 & 10.24& 5.51 & 5.23 & 5.15 & 6.79 & 6.32 \\
  Wing \cite{feng2018wing}  & - & 25M & -  & $\text{CVPR}_{2018}$                    & 4.99 & 8.43 & 5.21 & 4.88 & 5.26 & 6.21 & 5.81 \\
  MHHN \cite{wan2020robust} & 4 HG & - & - & $\text{TIP}_{2020}$                     & 4.77 & 9.31 & 4.79 & 4.72 & 4.59 & 6.17 & 5.82 \\
  DCFE \cite{valle2018deeply} & - & - & -  & $\text{ECCV}_{2018}$                    & 4.69 & 8.63 & 6.27 & 5.73 & 5.98 & 7.33 & 6.88 \\
  DeCaFA \cite{dapogny2019decafa} & - & $\sim$10M & -  & $\text{ICCV}_{2019}$        & 4.62 & 8.11 & 4.65 & 4.41 & 4.63 & 5.74 & 5.38 \\
  HRNet \cite{HRNET} & HRNetW18C  & 9.66M & 4.75G  & $\text{TPAMI}_{2020}$           & 4.60 & 7.94 & 4.85 & 4.55 & 4.29 & 5.44 & 5.42 \\
  AS w. SAN \cite{qian2019aggregation} & -  & 35.02M & 33.87G & $\text{ICCV}_{2019}$ & 4.39 & 8.42 & 4.68 & 4.24 & 4.37 & 5.60 & 4.86 \\
  LUVLI \cite{kumar2020luvli} & 8 HG  & - & - & $\text{CVPR}_{2020}$                 & 4.37 & 7.56 & 4.77 & 4.30 & 4.33 & 5.29 & 4.94 \\
  AWing \cite{wang2019adaptive} & 4 HG  & 24.15M & 26.8G & $\text{ICCV}_{2019}$      & 4.36 & 7.38 & 4.58 & 4.32 & 4.27 & 5.19 & 4.96  \\
  SDFL \cite{lin2021structure} & HRNetW18C & - & 5.17G & $\text{TIP}_{2021}$         & 4.35 & 7.42 & 4.63 & 4.29 & 4.22 & 5.19 & 5.08  \\
  SDL \cite{li2020structured} & HRNetW18C  & - & - & $\text{ECCV}_{2020}$            & 4.21 & 7.36 & 4.49 & 4.12 & 4.05 & {\color{blue}$\bm{4.98}$} & 4.82  \\
  ADNet \cite{huang2021adnet} & 4 HG &13.37M & 17.04G & $\text{ICCV}_{2021}$         & 4.14 & {\color{blue}$\bm{6.96}$} & 4.38 & 4.09 & 4.05 & 5.06 & 4.79  \\
  SLPT \cite{xia2022sparse} & HRNetW18C & 13.19M & 6.12G & $\text{CVPR}_{2022}$      & 4.14 & {\color{blue}$\bm{6.96}$} & 4.45 & {\color{blue}$\bm{4.05}$} & 4.00 & 5.06 & 4.79  \\
  SLPT$\uparrow$ \cite{xia2022sparse} & HRNetW18C & 19.45M & 8.14G & $\text{CVPR}_{2022}$ & {\color{blue}$\bm{4.12}$} & 6.99 & 4.37 & {\color{red}$\bm{4.02}$} & 4.03 & 5.01 & 4.79 \\

  \hline\hline
  $\text{HIH}_T$ & 1 HG          & 10.37M & 6.99G  & ours & 4.27 & 7.38 & 4.33 & 4.59 & 4.01 & 5.16 & 5.00 \\
  $\text{HIH}_S$ & 2 HG          & 14.47M & 10.38G & ours & 4.13 & 7.20 & {\color{blue}$\bm{4.18}$} & 4.38 & {\color{blue}$\bm{3.90}$} & {\color{blue}$\bm{4.98}$} & {\color{blue}$\bm{4.71}$} \\
  $\text{HIH}_B$ & 4 HG          & 22.68M & 17.15G & ours & {\color{red}$\bm{4.08}$} & {\color{red}$\bm{6.87}$} & {\color{red}$\bm{4.06}$} & 4.34 & {\color{red}$\bm{3.85}$} & {\color{red}$\bm{4.85}$} & {\color{red}$\bm{4.66}$}  \\
  \hline
  \end{tabular}
  \end{center}
  \caption{Comparison with state-of-the-art methods on WFLW with inter-ocular NME. Key: [{\color{red} \textbf{Best}}, {\color{blue} \textbf{Second Best}}, SLPT$\uparrow$ is with 12 layer]}
  \label{tab1_WFLW}
  \end{table*}

  \section{Experiments}
  \subsection{Experiments Settings} 
  \subsubsection{Datasets} 

  \emph{WFLW} dataset is a challenging one, which contains 7500 faces for training and 2500 faces for testing, based on WIDER Face\cite{yang2016wider} with 98 manually annotated landmarks. 
  The faces in WFLW introduce large variations in pose, expression, and occlusion. The testing set is further divided into six subsets for a detailed evaluation, 
  namely, pose (326 images), expression (314 images), illumination (698 images), make-up (206 images), occlusion (736 images) and blur (773 images).
  
  \emph{300W} dataset provides 68 landmarks for each face, where the face images are collected from LFPW\cite{belhumeur2013localizing}, AFW\cite{zhu2012face}, HELEN\cite{le2012interactive}, XM2VTS\cite{messer1999xm2vtsdb} and IBUG. 
  Following the protocol used in \cite{zhu2012face}, all 3148 training images are from the training set of LFPW and HELEN, and the full set of AFW. 
  The 689 testing images are further divided into three sets: the test samples (554 images) from LFPW and HELEN as the common subset, 
  the 135-image IBUG as the challenging subset, and the union of them as the full set.

  \emph{COFW} dataset consists of 1345 images for training and 507 faces for testing, where the face images have large pose variations, heavy occlusion and expression variations. 
  29 landmarks are provided for each face.
  
  \subsubsection{Evaluation Metrics}
  
  Following previous works\cite{HRNET,kumar2018disentangling,LABWFLW}, we use the standard metrics \emph{Normalized Mean Error}(NME), \emph{Area Under the Curve}(AUC) and \emph{Failure Rate}(FR), which are the most authoritative for face alignment.
  In each table, we report results using the same metric adopted in respective baselines.
  
  \emph{Normalized Mean Error (NME)} is defined as:
  
  \begin{equation}
     NME(\%) = \frac{1}{N} \sum_{k = 1}^{N} \frac{||P_k-\hat{P_k}||_2}{d}
  \end{equation}
  
  For the 300W and WFLW, the NME normalized by the inter-ocular distance is used. For the COFW, the NME normalized by inter-ocular distance and inter-pupil distance the are both adopted.
  $P_k \text{ and } \hat{P_k}$ denote the ground-truth and prediction location of the k-th landmark.
  For simplicity, we omit the \% symbol. 
  Lower NME is better to facial landmark detector.
  
  \emph{Area Under the Curve (AUC)} is calculated based on the cumulative error distribution (CED) curve.
  The AUC for a testset is computed as the area under the curve, up to the cut off NME value.
  Higher AUC represents better detector.
  
  \emph{Failure Rate (FR)} is another metric to evaluate localization quality, which refers to the percentage of images in the testset when NME is larger than a perfixed threshold.
  The Lower FR is corresponding to the better performance.

  \begin{table}[t!]
    \resizebox{0.47\textwidth}{!}{
      \centering
    \begin{tabular}{m{1.9cm}<{\centering}|m{1.2cm}<{\centering}|m{0.95cm}<{\centering}m{0.95cm}<{\centering}|m{0.95cm}<{\centering}m{0.95cm}<{\centering}}
      \hline
      \multirow{2}{*}{Method} & \multirow{2}{*}{Year} &  \multicolumn{2}{c|}{Inter-Ocular} & \multicolumn{2}{c}{Inter-Pupil} \\
      && NME(\%)$\downarrow$ & FR(\%)$\downarrow$ & NME(\%)$\downarrow$ & FR(\%)$\downarrow$ \\  \hline
      LAB \cite{LABWFLW} & $\text{CVPR}_{2018}$& 3.92 & 0.39 & 5.58 & 2.76 \\
      SDFL \cite{lin2021structure} &$\text{TIP}_{2021}$ & 3.63 & {\color{red}$\bm{0.00}$} & - & - \\
      HRNet \cite{HRNET} & $\text{TPAMI}_{2020}$& 3.45 & 0.20 & - & -\\

      DAC-CSR \cite{feng2017dynamic} & $\text{CVPR}_{2017}$& - & - & 6.03 & 4.73 \\
      Human \cite{COFW} & $\text{ICCV}_{2013}$ & -& - & 5.60 & - \\
      Wing \cite{feng2018wing} &$\text{CVPR}_{2018}$ &-&- & 5.44 & 3.75 \\
      DCFE \cite{valle2018deeply} & $\text{ECCV}_{2018}$ &-&-& 5.27 & 7.29 \\
      MHHN \cite{wan2020robust} &$\text{TIP}_{2020}$ &-&-& 4.95 & 1.78 \\
      AWing \cite{wang2019adaptive} & $\text{ICCV}_{2019}$ &-&-& 4.94 & 0.99 \\
      ADNet \cite{huang2021adnet} & $\text{ICCV}_{2021}$ &-&-& {\color{blue}$\bm{4.68}$} & {\color{blue}$\bm{0.59}$} \\
      SHN-GCN  \cite{zhang2020robust} & $\text{TIP}_{2020}$ &-&-& 5.67 & 2.36 \\
      SLPT \cite{xia2022sparse} &$\text{CVPR}_{2022}$ & 3.32 & {\color{red}$\bm{0.00}$} & 4.79 & 1.18 \\ 
      
      \hline\hline		
  
      $\text{HIH}_T$ & ours & 3.35 & {\color{red}$\bm{0.00}$} & 4.83 & {\color{blue}$\bm{0.59}$}  \\
      $\text{HIH}_S$ & ours & {\color{blue}$\bm{3.25}$} & {\color{red}$\bm{0.00}$} & 4.69 & {\color{blue}$\bm{0.59}$} \\
      $\text{HIH}_B$ & ours & {\color{red}$\bm{3.21}$} & {\color{red}$\bm{0.00}$}  & {\color{red}$\bm{4.63}$} & {\color{red}$\bm{0.39}$} \\
  
      \hline
    \end{tabular}
    }
    \caption{NME and FR$_{0.1}$ comparisons with state-of-the-art methods under Inter-Ocular normalization and Inter-Pupil normalization on COFW. [{\color{red} \textbf{Best}}, {\color{blue} \textbf{Second Best}}]}
    \label{tab1_cofw_sota}
  \end{table}
  
  \begin{table}[t]
  \begin{center}
  \begin{tabular}{l|c|ccc}
  \hline
  Method & Year & \tabincell{c}{ Full } & \tabincell{c}{Com.} & \tabincell{c}{ Chal. } \\
  \hline\hline
  LAB \cite{LABWFLW}  & $\text{CVPR}_{2018}$                  & 3.49 & 2.98 & 5.19  \\
  Wing \cite{feng2018wing}   & $\text{CVPR}_{2018}$           & 3.60 & 3.01 & 6.01  \\
  DCFE \cite{valle2018deeply}  & $\text{ECCV}_{2018}$         & 3.24 & 2.76 & 5.22  \\
  DeCaFA \cite{dapogny2019decafa}  & $\text{ICCV}_{2019}$     & 3.39 & 2.93 & 5.26 \\
  AS w. SAN \cite{qian2019aggregation}  & $\text{ICCV}_{2019}$& 3.86 & 3.21 & 6.49  \\
  AWing \cite{wang2019adaptive}  & $\text{ICCV}_{2019}$       & {\color{blue}$\bm{3.07}$} & 2.72 & {\color{red}$\bm{4.52}$} \\
  HRNet \cite{HRNET}  & $\text{TPAMI}_{2020}$                 & 3.32 & 2.87 & 5.15 \\
  LUVLI \cite{kumar2020luvli}  & $\text{CVPR}_{2020}$         & 3.23 & 2.76 & 5.16  \\
  SHN-GCN \cite{zhang2020robust} & $\text{TIP}_{2020}$        & 3.10 & 2.73 & 4.64 \\
  SDFL \cite{lin2021structure}  & $\text{TIP}_{2021}$         & 3.28 & 2.88 & 4.93 \\
  ADNet \cite{huang2021adnet}  & $\text{ICCV}_{2021}$         &  {\color{red}$\bm{2.93}$} & {\color{red}$\bm{2.53}$} &  {\color{blue}$\bm{4.58}$} \\
  SLPT \cite{xia2022sparse}  & $\text{CVPR}_{2022}$           & 3.17 & 2.75 & 4.90 \\
  
  \hline\hline
  $\text{HIH}_T$ & ours & 3.29 & 2.89 & 4.95  \\
  $\text{HIH}_S$ & ours & 3.15 & 2.75 & 4.80  \\
  $\text{HIH}_B$ & ours & 3.09 & {\color{blue}$\bm{2.65}$} & 4.89  \\
  
  \hline
  \end{tabular}
  \end{center}
  \caption{Comparison with state-of-the-art methods under inter-ocular NME on 300W.Key:[{\color{red} \textbf{Best}}, {\color{blue} \textbf{Second Best}}]}
  \label{tab1_300W_sota}
  \end{table}

  \begin{table}[!tbh]
  \begin{center}
  \resizebox{0.5\textwidth}{!}{
  \begin{tabular}{c|l|c|c|c|c|c|c|c}
  \hline
    Metric & Method & Fullset & \tabincell{c}{Pose} & \tabincell{c}{Exp.} & \tabincell{c}{Ill.} & \tabincell{c}{Mu.} & \tabincell{c}{Occ.} & \tabincell{c}{Blur} \\
    \hline\hline
    \multirow{12}{*}{$\text{FR}_{10}(\downarrow)$} 
      & LAB & 7.56 & 28.83 & 6.37 & 6.73 & 7.77 & 13.72 & 10.74 \\
      & SAN & 6.32 & 27.91 & 7.01 & 4.87 & 6.31 & 11.28 & 6.60 \\
      & HRNet & 4.64 & 23.01 & 3.50 & 4.72 & 2.43 & 8.29 & 6.34 \\
      & AS w. SAN & 4.08 & 18.10 & 4.46 & 2.72 & 4.37 &7.74 & 4.40\\
      & LUVLi & 3.12 & 15.95 & 3.18 &  {\color{blue}$\bm{2.15}$} & 3.40 & 6.39 &  {\color{blue}$\bm{3.23}$} \\
      & AWing & 2.84 & 13.50 & 2.23 & 2.58 & 2.91 & 5.98 & 3.75 \\
      & SDFL &  {\color{blue}$\bm{2.72}$} & 12.88 &  {\color{blue}$\bm{1.59}$} & 2.58 & 2.43 &  5.71 & 3.62 \\ 
      & SDL & 3.04 & 15.95 & 2.86 & 2.72 &  {\color{red}$\bm{1.45}$} &  {\color{blue}$\bm{5.29}$} & 4.01 \\
      & ADNet &  {\color{blue}$\bm{2.72}$} &  12.72 &  2.15 &  2.44 & {\color{blue}$\bm{1.94}$} & 5.79 & 3.54\\
      & SLPT &  2.76 &  {\color{blue}$\bm{12.27}$} & 2.23 &  {\color{red}$\bm{1.86}$} & 3.40 & 5.98 & 3.88 \\
      & SLPT$\uparrow$ & {\color{blue}$\bm{2.72}$} & {\color{red}$\bm{11.96}$} & {\color{blue}$\bm{1.59}$} & {\color{blue}$\bm{2.15}$} & {\color{blue}$\bm{1.94}$} & 5.70 & 3.88 \\
    \cline{2-9}
    ~&$\text{HIH}_T$           & 2.92 & 13.80 & 2.54 & 2.43 & 3.39 & 6.38 & 3.62 \\
    ~&$\text{HIH}_S$           & 2.80 & 13.49 & {\color{blue}$\bm{1.59}$} & 2.57 & {\color{blue}$\bm{1.94}$} & {\color{blue}$\bm{5.29}$} & {\color{blue}$\bm{3.23}$} \\
    ~&$\text{HIH}_B$           & {\color{red}$\bm{2.60}$} & 12.88 & {\color{red}$\bm{1.27}$} & 2.43 & {\color{red}$\bm{1.45}$} & {\color{red}$\bm{5.16}$} & {\color{red}$\bm{3.10}$} \\
    
    \hline
    \multirow{12}{*}{$\text{AUC}_{10}(\uparrow)$} 
    & LAB & 0.532 & 0.235 & 0.495 & 0.543 & 0.539 & 0.449 & 0.463 \\
    & SAN & 0.536 & 0.236 & 0.462 & 0.555 & 0.522 & 0.456 & 0.493 \\
    & HRNet & 0.524 & 0.251 & 0.510 & 0.533 & 0.545 & 0.459 & 0.452 \\
    & AS w. SAN & 0.591 & 0.311 & 0.549 & {\color{blue}$\bm{0.609}$} & 0.581 & 0.516 & 0.551 \\
    & LUVLi & 0.557 & 0.310 & 0.549 & 0.584 & 0.588 & 0.505 & 0.525 \\
    & AWing & 0.572 & 0.312 & 0.515 & 0.578 & 0.572 & 0.502 & 0.512 \\
    & SDFL & 0.576 & 0.315 & 0.550 & 0.585 & 0.583 & 0.504 & 0.515 \\ 
    & SDL &  0.589 & 0.315 & 0.566 & 0.595 & 0.604 & 0.524 & 0.533 \\ 
    & ADNet & {\color{blue}$\bm{0.602}$} & 0.344 & 0.523 & 0.580 & 0.601 & 0.530 & 0.548 \\ 
    & SLPT & 0.595 & 0.348 & 0.574 & 0.601 & 0.605 & 0.515 & 0.535 \\ 
    & SLPT$\uparrow$ & 0.596 & {\color{blue}$\bm{0.349}$} & 0.573 & 0.603 & 0.608 & 0.520 & 0.537\\
    \cline{2-9}
    ~&$\text{HIH}_T$           & 0.593 & 0.337 & 0.578 & 0.606 & 0.600 & 0.524 & 0.545 \\
    ~&$\text{HIH}_S$           & {\color{blue}$\bm{0.602}$} & 0.346 & {\color{blue}$\bm{0.592}$} & {\color{red}$\bm{0.613}$} & {\color{blue}$\bm{0.612}$} & {\color{blue}$\bm{0.535}$} & {\color{blue}$\bm{0.558}$} \\
    ~&$\text{HIH}_B$           & {\color{red}$\bm{0.605}$} & {\color{red}$\bm{0.358}$} & {\color{red}$\bm{0.601}$} & {\color{red}$\bm{0.613}$} & {\color{red}$\bm{0.618}$} & {\color{red}$\bm{0.539}$} & {\color{red}$\bm{0.561}$}\\
    \hline

  \end{tabular}
  }
  \end{center}
  \caption{Comparisons with state-of-the-art methods in AUC and FR on WFLW. Key: [{\color{red} \textbf{Best}}, {\color{blue} \textbf{Second Best}}, SLPT$\uparrow$ is with 12 layer]}
  \label{tab2_WFLW}
  \end{table}

  \begin{table}[!tbhp]
  \begin{center}
  \resizebox{0.5\textwidth}{!}{
  \begin{tabular}{c|l|ccccccc}
  \hline
  Metric & Method & Fullset & \tabincell{c}{Pose} & \tabincell{c}{Exp.} & \tabincell{c}{Ill.} & \tabincell{c}{Mu.} & \tabincell{c}{Occ.} & \tabincell{c}{Blur} \\
  \hline\hline
  \multirow{6}{*}{$\text{NME}(\downarrow)$} 
    & Baseline\cite{newell2016stacked}       & 4.53  & 7.91 & 4.72  & 4.74  & 4.32  & 5.48  & 5.23  \\
  ~ & WSM\cite{HRNET}             & 4.35  & 7.68 & 4.54  & \textbf{4.56}  & 4.14  & 5.33  & 5.07  \\
  ~ & DARK\cite{zhang2020distribution}            & 4.33  & 7.45 & 4.44  & 4.65  & 4.11  & 5.21  & 5.02  \\
  ~ & WOV\cite{xia2022sparse}             & 4.34  & 7.50 & 4.50  & \textbf{4.56}  & 4.12  & 5.17  & \textbf{4.98}  \\
  ~ & WOM\cite{law2018cornernet}             & 4.44  & 7.83 & 4.61  & 4.79  & 4.15  & 5.33  & 5.20  \\
  ~ & HIH(ours)             & \textbf{4.27} & \textbf{7.38} & \textbf{4.33} & 4.59 & \textbf{4.01} & \textbf{5.16} & 5.00 \\
  \hline
  
  \hline
  \multirow{6}{*}{$\text{FR}_{10}(\downarrow)$} 
    & Baseline\cite{newell2016stacked}       & 4.00  & 19.02 & 3.82  & 3.15  & 3.88  & 7.74  & 4.27  \\
  ~ & WSM\cite{HRNET}             & 3.64  & 18.10 & 3.18  & 3.01  & 3.88  & 7.61  & 3.62  \\
  ~ & DARK\cite{zhang2020distribution}            & 3.16  & 16.87 & \textbf{1.91}  & 2.58  & 2.91  & 6.52  & \textbf{3.49}  \\
  ~ & WOV\cite{xia2022sparse}             & 3.20  & 15.33 & 2.55  & 2.58  & \textbf{1.94}  & 6.39  & 3.88  \\
  ~ & WOM\cite{law2018cornernet}             & 3.28  & 15.95 & 2.55  & 3.01  & 2.91  & \textbf{6.25}  & 4.27  \\
  ~ & HIH(ours)             & \textbf{2.92} & \textbf{13.80} & 2.54 & \textbf{2.43} & 3.39 & 6.38 & 3.62 \\
  \hline
  
  \hline
  \multirow{6}{*}{$\text{AUC}_{10}(\uparrow)$} 
    & Baseline\cite{newell2016stacked}       & 0.568  & 0.291 & 0.547  & 0.579  & 0.574  & 0.497  & 0.516  \\
  ~ & WSM\cite{HRNET}             & 0.586  & 0.310 & 0.565  & 0.596  & 0.591  & 0.511  & 0.531  \\
  ~ & DARK\cite{zhang2020distribution}            & 0.587  & 0.328 & 0.571  & 0.599  & 0.594  & 0.518  & 0.539  \\
  ~ & WOV\cite{xia2022sparse}             & 0.585  & 0.320 & 0.570  & 0.594  & 0.593  & 0.517  & 0.538  \\
  ~ & WOM\cite{law2018cornernet}             & 0.579  & 0.308 & 0.559  & 0.586  & 0.590  & 0.507  & 0.526  \\
  ~ & HIH(ours)             & \textbf{0.593} & \textbf{0.337} & \textbf{0.578} & \textbf{0.606} & \textbf{0.600} & \textbf{0.524} & \textbf{0.545} \\
  \hline

  \end{tabular}
  }
  \end{center}
  \caption{Comparisons with solutions under NME, AUC and FR on WFLW.}
  \label{tab2_WFLW_solution}
  \end{table}

  \begin{table}[t]
  \begin{center}
  \begin{tabular}{l|c|cccc}
  \hline
  Method & Flops & \tabincell{c}{ Val} & \tabincell{c}{Com.} & \tabincell{c}{ Cha. }  & \tabincell{c}{ Test. }\\
  \hline\hline
  Baseline & 6.67G & 3.581 & 3.157 & 5.321 & 4.132  \\
  WSM  &  6.67G  & 3.420 & 2.988 & 5.193 & 3.999 \\
  DARK &  6.87G  & 3.404 & 2.962 & 5.221 & 3.908 \\
  WOV  &  6.91G  & 3.448 & 3.014 & 5.228 & 4.010 \\
  WOM  &  6.85G  & 3.640 & 3.127 & 5.745 & 4.445 \\
  \hline\hline
  $\text{HIH} \text{ w. } L_2$                  & 6.93G  & 3.356 & 2.946 & 5.038 & 3.835 \\
  $\text{HIH} \text{ w. } \mathcal{L}_{csc}$  & 6.93G  & \textbf{3.296} & \textbf{2.891} & \textbf{4.956} & \textbf{3.832} \\
  \hline
  \end{tabular}
  \end{center}
  \caption{Comparison with other solutions under NME on 300W and ablation study between MSE and coordinate soft-classification loss.}
  \label{tab2_300W_solution}
  \end{table}

  \subsubsection{Implementation Details}

  To produce the inputs of model, we crop face regions and resize them into $256 \times 256$,
  then the images are augmented by a set of data augmentations, i.e. horizontal flip ($50\%$), rotation ($\pm 30^{\circ },50\%$), occlusion ($50\%$), and Gaussian blur ($30\%$).
  And the cropped images of training or testing are resized to 256$\times$256 resolution in the preprocessing. 
  Following previous works' setting\cite{wang2019adaptive,kumar2020luvli}, we take Hourglass as the network backbone, and set 64$\times$64 for the resolution of integer heatmaps.
  We use L2 loss for the integer heatmap and set Gauss sigma 1.5 except 300W 1.0\cite{HRNET}.
  For decimal heatmap setting, we set $8\times8$ resolution, sigma equals 1.0.
  We denote $\text{HIH}_T$, $\text{HIH}_S$, and $\text{HIH}_B$ to represent tiny, small, and base HIH, which are composed of 1,2 and 4 stacks of HG, respectively.
  In $\text{HIH}$, we set Gauss sigma of decimal heatmap to 1.0, weight is 0.1 in $\text{HIH}_T$, and weight 0.05 in $\text{HIH}_S$ and $\text{HIH}_B$.
  All experiments are conducted on the RTX 2080Ti device without pretraining.
  And the batch size is set to 16 for one GPU, all losses are with mean reduction.
  For the training procedure, we used the Adam \cite{kingma2014adam} optimizer with init learning rate 5e-6, which decreases 10 times at 30 and 50 epochs.
  And the total training epochs is 60, which is the same as previous works\cite{HRNET}.

  \subsection{Quantitative Result of Quantization Error}
\begin{figure}[tbh]
      \centering
      \newpage
      \includegraphics{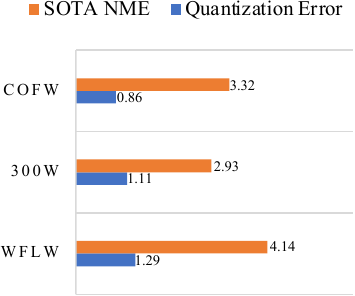}
      \caption{Inter-ocular NME distribution on benchmarks, the blue parts represent the NME by quantization error under ideal conditions, and the orange parts represent state-of-the-art methods\cite{huang2021adnet,xia2022sparse} caused. 
      }
      \label{fig:bottleneck}
    \end{figure}

    We have conducted experiments to explore how much impact the quantization error caused.
    We directly take the ground-truth integer heatmaps as the prediction (loss equals zero) to decode the sub-optimal coordinates, and calculate the NME, which is caused by quantization error.
    It is tested with 256x256 input resolution and 64x64 heatmap resolution on three benchmarks, and here we take the result of ADNet\cite{huang2021adnet} or SLPT\cite{xia2022sparse} as SOTA item.
    As shown in Fig.\ref{fig:bottleneck}, NME generated by quantization error is 0.864 on COFW(29 pts), 1.111 on 300W(68 pts), and 1.285 on WFLW(98 pts), even larger than 1/3 of the state-of-the-art methods caused NME.
    With the increasing trend of the number of keypoint in the dataset, the corresponding error will continue to rise, revealing the importance of this issue.

    \subsection{comparison with State-of-the-art Methods}
  
  To compare our networks against the state-of-the-art methods, we conduct evaluations on various benchmarks.

  \textbf{Comparison on WFLW}: Following their evaluation protocol\cite{LABWFLW}, we report results in terms of NME, AUC, and FR by inter-ocular normalization.
  As tabulated in Tab.\ref{tab1_WFLW} and Tab.\ref{tab2_WFLW},  HIH demonstrates  impressive performance.
  With increasing the hourglass, the performance of HIH can be further improved and outperforms the state-of-the-art methods, such as ADNet and SLPT.
  Our tiny and small networks also perform remarkable result with light-weight parameters and flops,
  Specifically,  our $\text{HIH}_S$  reaches 4.13 \text{NME}, 2.80 $\text{FR}_{10}$, and 0.602 $\text{AUC}_{10}$.
  $\text{HIH}_B$ even reaches \textbf{4.08} NME and \textbf{2.60} $\text{FR}_{10}$  0.605 $\text{AUC}_{10}$  on WFLW,
  which demonstrates that our method could localize the landmarks accurately.

  \textbf{Comparison on COFW}: Following previous works, we report results in terms of NME and FR by inter-ocular normalization and inter-pupil normalization.
  In this experiment, the number of the training sample is relatively small, which leads to the apparent degradation of previous methods, such as SDFL and AWing.
  As shown in Tab.\ref{tab1_cofw_sota}, nevertheless, HIH still maintains excellent performance and shows superiority over all state-of-the-art methods.
  Significantly, even $\text{HIH}_S$ achieves a better performance than SLPT and ADNet.
  Furthermore, $\text{HIH}_B$ reaches \textbf{3.21} inter-ocular NME, 4.63 inter-pupil NME and failure rate decreases to 0.39.

  \textbf{Comparison on 300W}:
  It also depicts the comparison in inter-ocular NME on the 300W benchmark that contains full set, challenge set, and common set.
  Compared with previous works, HIH also shows impressive performance. 
  Especially, $\text{HIH}_B$ reaches 3.09 on the full set, 2.65 on common set and 4.90 on challenge set.
  With limited training samples, the methods with prior knowledge, such as facial boundaries (AWing and ADNet) 
  , achieve better performance.
  Compared with other methods, HIH performs better than LUVLI, SDFL, and SLPT.

  \subsection{comparison with Other Solutions}
  \label{sec:comparison_solutions}
  
  Here we discuss the result in comparison with other works in addressing quantization error,
  which is divided into face alignment and other fields.
  In the literature of face alignment, these approaches include SHN-GCN\cite{zhang2020robust}, FHR\cite{tai2019towards}, MHHN\cite{wan2020robust}, LUVLI\cite{kumar2020luvli}, ADNet\cite{huang2021adnet}, SLPT\cite{xia2022sparse}.
  Besides, HRNet's practice leverages the neighbor points to calculate the gradient as the offset.
  As shown in Fig.\ref{tab1_WFLW}, except SLPT and HRNet taking HRNetW18C as the backbone, the methods all take 4-stacked Hourglass as the backbone.
  And as shown in previous discussions, our HIH obviously performs better among them.
  Therefore, to further demonstrate the feasibility of HIH, we here discuss the combination between hourglass and solutions of SLPT as well as HRNet.
  Besides, we also combine classic and state-of-the-art solution from pose estimation and object detection, the detailed design with the above solutions are discussed in Appendix\ref{sec:other_solution}, which includes With Offset Value(WOV), With Offset Map(WOM), DARK, With Second Maximal(WSM).
  About baseline without recovering subpixel part, we expand the channel from 256 to 280, which is used to counteract the effect of increasing the recovery part.

  \begin{figure}[tbh]
    \centering
    \newpage
    \includegraphics[width=0.95\linewidth]{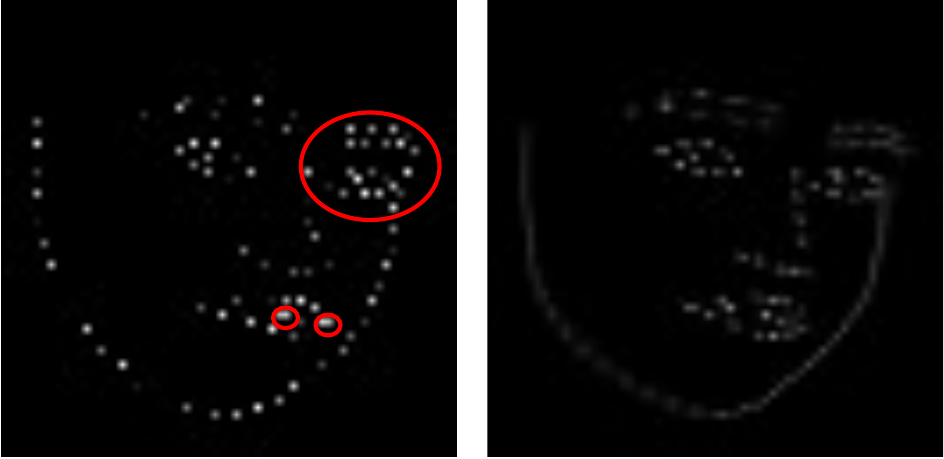}
    \caption{Comparison between ground-truth and predicted, in WOM's x-offset map. left is the ground-truth, right is predicted. There are spatial location conflicts in the red area.
    The predicted offset heatmap generates continuous boundary lines instead of discrete points, with unclear semantic information.
    }
    \label{wom_difference}
  \end{figure}

  Tab.\ref{tab2_WFLW_solution} shows the comparison results under NME, AUC, and FR metrics on WFLW dataset under 1 HG. 
  The results demonstrate that all the five solutions  achieve positive results, showing that quantization error exists indeed.
  Among them, WOV achieves very close performance to WSM with only a slight gap, while WOM shows the lowest improvements. 
  Our HIH performs better than all the other methods, which demonstrates HIH representation outperforms other methods.   
  Tab.\ref{tab2_300W_solution} shows the comparisons in NME and the parameter information on 300W with $\text{HIH}_T$. 
  Note that WSM recovers the quantization error based on the original heatmap, so it has the same parameters and flops as the baseline.
  From the table, WSM, WOV, and our HIH have a positive gain effect, while WOM does not.
  Furthermore, HIH shows significantly higher improvements than the others again, and  WSM is slightly better than WOV.
  
  We also discuss the effect of WOM as shown in Fig.\ref{wom_difference} to emphasize the difference between HIH and WOM.
  The interval among facial landmarks is much closer than the interval among objects from the object detection field, 
  as well as the larger amount of points, which makes points highly dense.
  After downsampling, the aforementioned differences cause the probability of occupancy conflict to be higher, as shown in Fig.\ref{wom_difference}(a) red circles.
  Especially in the large-pose situation, these points are very closed or even overlapped, and the network utilizes the SmoothL1 loss function to constrain the offset map \cite{law2018cornernet}.
  As shown in Fig.\ref{wom_difference}(b), these factors cause predictions to be a continuous contour map instead of discrete points, and the activation of the corresponding location is not prominent enough.
  

  \subsection{Comparison between Soft Label and Hard Label}

  It is worth noting again the difference and feasibility between the soft label and hard label.
  The hard label only sets the value on the location which is with the maximal response, and soft labels also set values to nearby locations.
  These soft labels provide more positive samples than the hard label, which leverages the spatial support and enlarges practical constraints.

  We have carried out experiments on using the hard label to represent the decimal value, and the result is shown in Tab.\ref{hard-soft label}.
  The performance of soft labels is superior to a significant margin over the hard representation on the WFLW benchmark \cite{LABWFLW}, which demonstrates that more positive samples are needed in supervising the sparse map.

  We also have tried to combine the classification task with the original integer heatmap, but it failed to classify numerous labels due to the resolution is $64 \times 64$.
  If the sigma is set small (1,2), there are very few non-zero values on the heatmap, resulting in too many background labels for classification.
  If the sigma is set large, the ambiguity of the generated heatmap is relatively large, and the keypoint should not have the non-zero response at this position, resulting in obscurity in the classification and little difference among the probability values.
  Therefore, for such large-resolution heatmap regression, it is not appropriate to convert it to a classification task.

  \begin{table}[t]
  \begin{center}
  \resizebox{0.47\textwidth}{!}{
  \begin{tabular}{c|c|cccccccccc}
  \hline
  Design & network & Fullset & \tabincell{c}{Pose} & \tabincell{c}{Exp.} & \tabincell{c}{Ill.} & \tabincell{c}{Mu.} & \tabincell{c}{Occ.} & \tabincell{c}{Blur} \\
  \hline\hline
  Hard   & $\text{HIH}_T$ & 4.96 & 8.29 & 5.19 & 5.19 & 4.65 & 5.78 & 5.57 \\
  Soft   & $\text{HIH}_T$ & 4.27 & 7.38 & 4.33 & 4.59 & 4.01 & 5.16 & 5.00 \\
  \hline
  Hard   & $\text{HIH}_S$ & 4.79 & 7.89 & 4.93 & 5.09 & 4.56 & 5.59 & 5.38 \\
  Soft   & $\text{HIH}_S$ & 4.13 & 7.20 & 4.18 & 4.38 & 3.90 & 4.98 & 4.71 \\
  \hline
  \end{tabular}
  }
  \end{center}
  \caption{Comparisons about hard and soft representation in the decimal heatmap.}
  \label{hard-soft label}
  \end{table}

  \subsection{Ablation Study}
  Our method consists of two pivotal components, $i.e.$ representation method, and loss function.
  Furthermore, the representation method includes decimal heatmap setting (resolution, sigma) and architecture design.
  To highlight the independent advantage, we further carry out the ablation study on these parts.

\begin{table}[t]
\begin{center}
\resizebox{0.48\textwidth}{!}{
\begin{tabular}{c|c|ccccccc}
\hline
Loss & network & Fullset & \tabincell{c}{Pose} & \tabincell{c}{Exp.} & \tabincell{c}{Ill.} & \tabincell{c}{Mu.} & \tabincell{c}{Occ.} & \tabincell{c}{Blur} \\
\hline\hline
MSE                   & $\text{HIH}_T$ & 4.31 & 7.40 & 4.36 & 4.52 & 4.08 & 5.17 & 5.00 \\
$\mathcal{L}_{csc}$   & $\text{HIH}_T$ & 4.27 & 7.38 & 4.33 & 4.59 & 4.01 & 5.16 & 5.00 \\
\hline
MSE                   & $\text{HIH}_S$ & 4.18 & 7.20 & 4.19 & 4.45 & 3.97 & 5.00 & 4.81  \\
$\mathcal{L}_{csc}$   & $\text{HIH}_S$ & 4.13 & 7.20 & 4.18 & 4.38 & 3.90 & 4.98 & 4.71 \\
\hline
MSE                   & $\text{HIH}_B$ & 4.12 & 6.99 & 4.20 & 4.45 & 3.93 & 4.88 & 4.71  \\
$\mathcal{L}_{csc}$   & $\text{HIH}_B$ & \textbf{4.08} & \textbf{6.87} & \textbf{4.06} & \textbf{4.34} & \textbf{3.85} & \textbf{4.85} & \textbf{4.66} \\
\hline
\end{tabular}
}
\end{center}
\caption{Ablation Study: Comparisons about decimal's loss function in the NME on the WFLW. }
\label{ablation_loss}
\end{table}



  \subsubsection{Comparison between ${L}_{csc}$ and MSE}
In this subsection, we investigate the effect of the proposed loss  ${L}_{csc}$. 
We replace ${L}_{csc}$ with the MSE loss and make a comparison on WFLW and 300W.
As shown in the Tab.\ref{ablation_loss}, $\mathcal{L}_{csc}$ achieves a noticeable improvement on all three networks, which also indicates the classification loss is more suitable for decimal heatmaps. 
Tab.\ref{tab2_300W_solution} also shows the superiority of $\mathcal{L}_{csc}$ than MSE on 300W dataset. 
Besides, our HIH with MSE loss also achieves better performance than SOTA, which demonstrates that our heatmap-in-heatmap design is feasible.
  \subsubsection{Comparison with different resolution}
  
  \begin{table}[t]
  \begin{center}
  \resizebox{0.48\textwidth}{!}{
  \begin{tabular}{c|c|cccccccccc}
  \hline
  Resolution & Ideal & Fullset & \tabincell{c}{Pose} & \tabincell{c}{Exp.} & \tabincell{c}{Ill.} & \tabincell{c}{Mu.} & \tabincell{c}{Occ.} & \tabincell{c}{Blur} \\
  \hline\hline
  4x4   & 0.420 & 4.389  & 7.553 & 4.471  & 4.613  & 4.152  & 5.201  & 5.023  \\
  8x8   & 0.182 & \textbf{4.273}  & \textbf{7.388} & \textbf{4.330}  & 4.595  & \textbf{4.005}  & \textbf{5.158}  & \textbf{4.997} \\
  16x16 & \textbf{0.091} & 4.307  & 7.438 & 4.337  & \textbf{4.576}  & 4.051  & 5.198  & 5.029  \\
  \hline
  \end{tabular}
  }
  \end{center}
  \caption{Ablation Study: Comparisons about decimal heatmap's resolution in the NME on the WFLW. The ideal item represents the errors under ideal conditions.}
  \label{ablation_resolution}
  \end{table}
  
  We further carry out the ablation study on the decimal heatmap resolution from $4\times4$, $8\times8$ to $16\times16$.
  As shown in Tab.\ref{ablation_resolution}, the $8\times8$ achieves the best performance while $4\times4$ setting produces the worst.
  We suppose the reason for the poor results of $4\times4$ is that applying Gaussian distribution by Eq.\ref{Gaussian distribution} on this small heatmap is inappropriate and causes too many large probability values.
  Although $16\times16$ setting makes a lower ideal error due to its high resolution, the experimental results under this setting are lower than $8\times8$ based results. 
  According to Proposition \ref{props}, our $8\times8$ decimal heatmap achieves lower coordinate representation precision than 1 pixel in the original images if the original image size is below or equal to $512\times512$.
  Since nearly all the face images are smaller than $512\times512$, $8\times8$ decimal heatmap is enough to cure the quantization errors, and the lower coordinate representation precision in $16\times16$ setting is point-less.
  Furthermore, with $16\times16$ decimal heatmaps, the semantic representation of heatmap pixels becomes less discriminative, and the increased amount of classification categories  makes network training more difficult. 
  
  
\begin{table}[t]
\begin{center}
\begin{tabular}{c|ccccccccccc}
\hline
Sigma  & Fullset & \tabincell{c}{Pose} & \tabincell{c}{Exp.} & \tabincell{c}{Ill.} & \tabincell{c}{Mu.} & \tabincell{c}{Occ.} & \tabincell{c}{Blur} \\
\hline\hline
0.5    & 4.21  & 7.05 & 4.34  & 4.54  & 4.04  & 4.90  & 4.73  \\
0.75   & 4.20  & 6.97 & 4.38  & 4.46  & 4.04  & 4.88  & 4.81  \\
1.0    & 4.08  & 6.87 & 4.06  & 4.34  & 3.85  & 4.85  & 4.66  \\
1.5    & 4.14  & 7.02 & 4.24  & 4.45  & 3.93  & 4.84  & 4.70  \\
\hline
\end{tabular}
\end{center}
\caption{Ablation Study: Comparisons about decimal's sigma in the NME on the $\text{HIH}_B$, tested on WFLW benchmark.}
\label{ablation_sigma}
\end{table}

\subsubsection{The Super-Parameter $\sigma$}
The ground-truth generation involves the super-parameter $\sigma$ in Gaussian distribution, so we take the ablation study on $\sigma$ from 0.5 to 1.5.
As shown in the Tab.\ref{ablation_sigma}, our method achieves the best performance when $\sigma$=$1.0$, and we use it as the default value in the paper. The second-best performance comes by setting $\sigma$ to 1.5 while both $0.5$ and $0.75$ show alarming results. We assume that the generated heatmap with $\sigma$=$0.5$ or $0.75$ is very sparse, and it loses numerous soft labels with positive values. And few positive labels significantly increase the difficulty of training, which results in bad performance.

\section{Conclusion}

In this paper, we quantitatively investigate the quantization error in face alignment and propose a novel representation method named HIH.
It uses a heatmap-in-heatmap scheme to represent the final coordinates jointly. 
Besides, we transform the subpixel coordinate regression into coordinate interval classification, and leverage the feature of the Gaussian distribution density function to design a seamless loss function, i.e., coordinate soft-classification loss, to learn the probability distribution in the decimal heatmaps. 
Extensive experiments show that our method is feasible in face alignment and outperforms other solutions.

{\appendix[Combination with Other Solutions]
\label{sec:other_solution}
In the literature of quantization error, the methods include Hourglass(with second maximal), DARK(Gaussian operation), CornerNet(with offset map), SLPT(with offset value), soft-argmax\cite{luvizon20182d,bulat2021subpixel}
HRNet(with near responses), G-RMI(vote) \cite{papandreou2017towards}, LUVLI(mean estimator)\cite{kumar2020luvli}, and so on.
Because HIH has compared with methods with 4-stacked Hourglass backbone, 
here we pass them and compare HIH with remaining solutions, which divided remaining methods into three categories according to their features.

\begin{figure}[!tbhp]
\begin{center}
\includegraphics[width=1.0\linewidth]{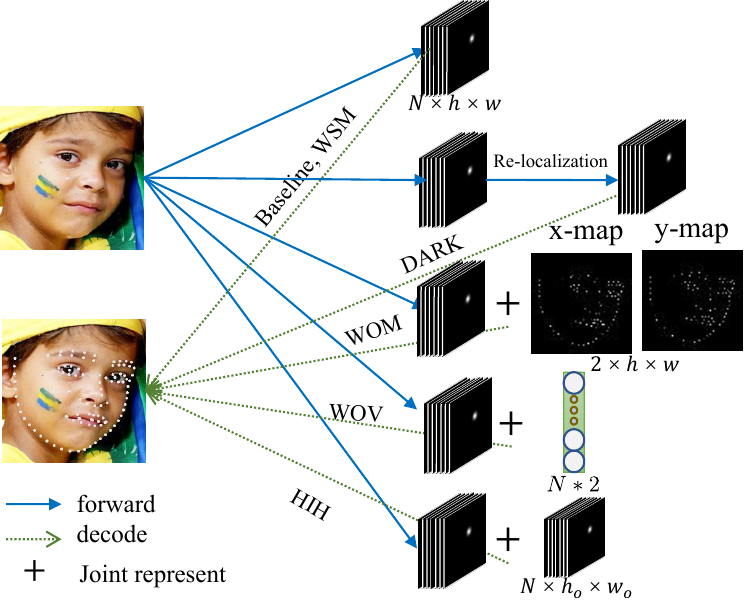}
\end{center}
\caption{Combinations with all designs. WSM is the commonly used practice, WOM is followed by Cornernet\cite{law2018cornernet}, Dark\cite{zhang2020distribution} is SOTA in pose estimation, WOV is inspired by \cite{zhang2020robust,xia2022sparse}, and HIH is ours.
  }
\label{tab:appendix_img}
\end{figure}

\subsection{With Near Response}
As described in the related work, there exist most works \cite{luvizon20182d,bulat2021subpixel,papandreou2017towards,kumar2020luvli,zhang2020distribution,wan2020robust,HRNET,newell2016stacked,tai2019towards} utilizing nearby location of maximal response to refine the accurate coordinates.
HRNet takes Hourglass' practice to calculate the location and gradient between first and second maximal responses, we noted with second maximal (WSM).
And in this direction, DARK is the state-of-the-art method presently, which is combined with Gaussian distribution to adjust the irregular heatmap.
To verify the effectiveness of HIH, here we take the common practice WSM, and the state-of-the-art work DARK \cite{zhang2020distribution} as comparisons.

\subsection{With Offset Map}
In the field of object detection, keypoint-based methods \cite{law2018cornernet,zhou2019bottom} introduce offset maps, which have the same resolution as heatmaps, to predict the offsets in $x$ and $y$ dimensions.
While each object in object-detection detects more than one point,  
one point is enough to locate a facial landmark, so we reduce the number of offset maps relatively.
Specifically, given a heatmap with size $K\times h\times w$, we use two offset maps with size $h\times w$, i.e., $\bm{O}^x$ and $\bm{O}^y$, to represent the $x$ and $y$ coordinates of offset values respectively.
Here $K$ is the number of landmarks. And
we first generate ground-truth offset $\bm{p}_k = (\frac{x_k}{n} - \lfloor \frac{x_k}{n} \rfloor ,  \frac{y_k}{n} - \lfloor \frac{y_k}{n} \rfloor )$. 
Here $n$ is the downsampling factor 
, $x_k \text{ and } y_k$ are the ground-truth coordinates of the $k$-th landmark. Thus the value of $\bm{O}^x$ in coordinate $\bm{q}_k$ is $O^x_{\bm{q}_k} = {\bm{p}_k}_x$ and  0 otherwise.
Here $\bm{q}_k=(x_k,y_k)$ is the $k$-th ground-truth landmark location, and ${\bm{p}_k}_x$ is the $x$ coordinate of the $\bm{p}_k$. $\bm{O}^y$ is also generated in the same way.
And the $k$-th predicted location is finally computed as
\begin{equation}
  \label{eq:addod}
  \bm{\hat{P}}_k = {\frac{(x^m, y^m)+\left(\bm{O}^x_{\hat{\bm{q}}_k},\bm{O}^y_{\hat{\bm{q}}_k}\right)}{(w,h)}} \ \ \ \ \ k = 1 \dots N
\end{equation}
where $\hat{\bm{q}}_k$=$(x^m, y^m)$ is the coordinate of predicted maximal response on the $k$-th heatmap.
$\bm{O}^x_{\bm{q}_k} \text{ and } \bm{O}^y_{\bm{q}_k}$ are the corresponding offset value of $\bm{O}^x \text{ and } \bm{O}^y$ in coordinate $\bm{q}_k$.
$(w,h)$ is the same normalized factor for the heatmap resolution.

\subsection{With Offset Value}
SLPT\cite{xia2022sparse} and SHN-GCN\cite{zhang2020robust} leverage coordinate regression and heatmap regression to represent the final location.
SLPT takes the Transformer structure and the last FC layer to regress the offset, while SHN-GCN takes the global constraint network to leverage multiple level features and last fc to regress the offset.
They both take heatmap to represent the integer coordinate, and regression value to represent the subpixel coordinate.
To make a fair comparison with previous work,  we still take CNN structure and final FC layer to regress the offset \cite{zhang2020robust}, and take the fusion of heatmap result and backbone feature as input of subpixel regression module\cite{xia2022sparse}.

The prediction is generated by combining heatmap results with direct offset results.
During network inference, the $k$-th landmark location is computed as following:



\begin{equation}
  \bm{\hat{P}}_k = \frac{(x^m, y^m) + \hat{\bm{p}}_k}{(w,h)}  \ \ \ \ \ k = 1 \dots N
\end{equation}
where $\hat{\bm{p}}_k$ is predicted offset value, others are the same as before.

}

\bibliographystyle{IEEEtran}
\bibliography{IEEEabrv,egbib}



 
%

\vfill

\end{document}